\documentclass[letterpaper,10pt,conference]{ieeeconf}

\IEEEoverridecommandlockouts
\usepackage{amsmath}
\usepackage{amssymb}
\usepackage{graphicx}
\usepackage{xcolor}
\usepackage{cite}

\title{\LARGE \bf
GLoTouch: Global-to-Local Haptic Perception Using a Parallel Gripper for
Object Search, Recognition, and Grasping Without External Vision
}

\author{
\authorblockN{
Zonglin Li\textsuperscript{1},
Wanruo Zhang\textsuperscript{2},
Yiming Wang\textsuperscript{2},
Kun Song\textsuperscript{3},
Xinyi Zhou\textsuperscript{2},
and Daolin Ma\textsuperscript{2,*}
}
\authorblockA{
\textsuperscript{1}Independent Researcher
\quad
\textsuperscript{2}Shanghai Jiao Tong University
\quad
\textsuperscript{3}The University of Hong Kong
}
}

\begin{document}

\maketitle
\thispagestyle{empty}
\pagestyle{empty}

\begin{abstract}
Perceiving objects in the environment is a fundamental capability of
autonomous robots. In dark or low-light environments, external cameras often
fail to reliably perceive object positions and geometry; when visual sensing
is unavailable, completing target search, recognition, and grasping through
touch alone becomes a key robot manipulation capability. This task must
simultaneously address container-scale spatial exploration and object-scale
fine-grained geometric perception, which is particularly challenging for
low-degree-of-freedom parallel grippers. However, a unified framework remains
lacking for connecting container-scale spatial exploration with object-scale
fine-grained geometric perception and grasping. To address this challenge, we
present \textbf{GLoTouch}, a global-to-local haptic perception and
manipulation framework built on a parallel gripper. In the global stage, the
gripper holds a passive long-reach probe, combining force measurements with
known tool geometry to localize contacts and actively estimate
candidate-object positions, coarse contours, and heights. In the local
stage, the robot sets down the probe and uses the bilateral visuotactile
sensors on the same gripper to directly acquire local haptic observations,
which are matched against a given target 3-D model without object-specific
training. We evaluate the framework in both simulation and real-robot
experiments. Source code will be open-sourced.
\end{abstract}

\section{INTRODUCTION}
\label{sec:introduction}

Robot manipulation typically relies on external vision for object
localization, recognition, and grasping. However, in dark warehouses,
nighttime cargo areas, or poorly illuminated container environments,
cameras may fail to capture usable images of the scene; object positions
and geometry therefore become unreliable. In such settings,
visual failure is not merely degraded perceptual accuracy but a boundary
condition for the entire manipulation pipeline. Haptic perception obtains
information through physical contact and does not depend on external
illumination, line of sight, or target appearance, providing a feasible
route to target retrieval without visual observation \cite{li2020review}.
This paper investigates how, given only the target object's 3-D model and
without external visual observation, a robot equipped with a parallel
gripper can use touch alone to search for, recognize, and retrieve the
corresponding target from a multi-object container.

Contact-based target retrieval nevertheless presents three difficulties.
First, haptic observations are inherently local: information becomes
available only through physical interaction, so the robot must actively
explore a container-scale workspace
\cite{li2020review,pai2023tactofind}. Second, every observation involves
force exchange; pushing or rotating a light, movable object may change its
pose, making previously acquired contacts difficult to reuse
\cite{zhong2022soft,suresh2021tactile}. Third, global search and local
recognition impose different sensing requirements: the former requires
long reach and sensitive collision detection, whereas the latter requires
stable contact and spatially resolved local geometry
\cite{pai2023tactofind,kamtikar2026tactful}. A parallel gripper provides
stable bilateral contact and simple control, well suited for precise local
recognition; extending this capability to container-scale search, however,
requires an additional sensing strategy.

Prior systems have demonstrated haptic search and retrieval using
multi-fingered hands and learned tactile policies, including TactoFind and
TACTFUL
\cite{pai2023tactofind,kamtikar2026tactful}.
Slender tools and tactile whiskers can extend contact range during global
exploration \cite{xiao2022active,zhao2024unknown}, while active tactile methods
select and aggregate local contacts for object recognition or geometric
reconstruction \cite{xu2023tandem3d,zhao2023fingerslam,comi2024touchsdf}.
These studies establish important foundations, but how to use a widely
available, low-degree-of-freedom parallel gripper to perform both large-scale
force-based search and high-resolution local tactile recognition remains
less explored. GLoTouch bridges these two scales by temporarily attaching a
probe to the gripper while retaining it for the final grasp.

\begin{figure*}[t]
    \centering
    \includegraphics[width=\linewidth]{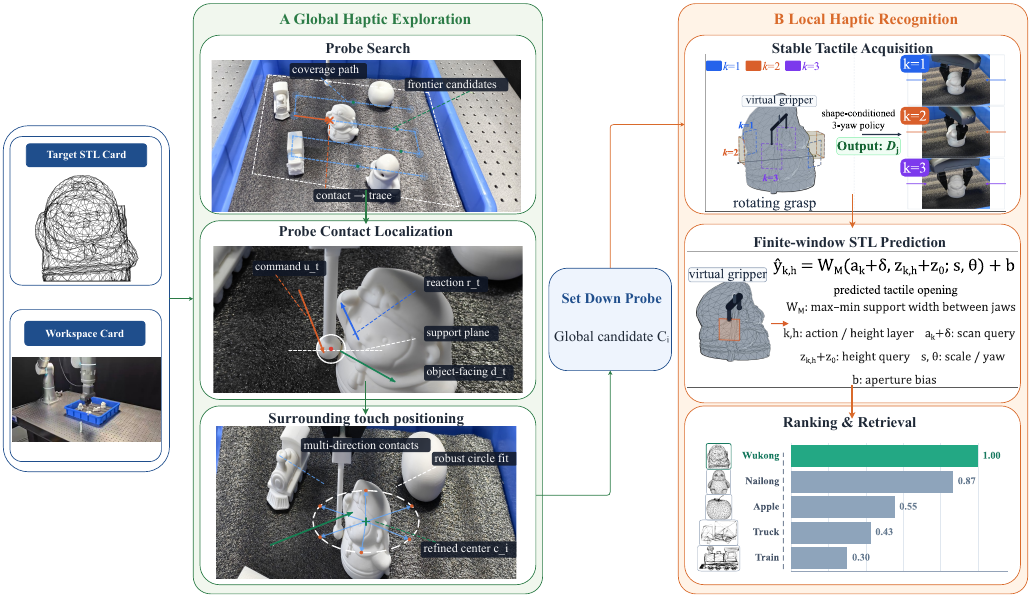}
    \caption{
        GLoTouch pipeline. Given the target STL model and workspace,
        \textbf{A} Global Haptic Exploration performs coverage-driven probe
        search, tool-geometry-aware contact localization, and surrounding
        touch positioning to estimate a candidate object $C_i$. After the
        probe is set down, \textbf{B} Local Haptic Recognition acquires
        stable tactile contacts, predicts finite-window support geometry
        from the target STL, and ranks candidates for retrieval. The
        ranking bars in \textbf{B} are illustrative similarity scores; the
        implemented decision ranks candidates by matching loss, where lower
        is better (Sec.~\ref{sec:local_recognition}).
    }
    \label{fig:glotouch_pipeline}
\end{figure*}

Specifically, \textbf{GLoTouch} is a global-to-local haptic framework built
around the same parallel gripper (Fig.~\ref{fig:glotouch_pipeline}). During
global search, a tool-geometry-aware
exploration procedure temporarily holds a passive long-reach probe: contact
measurements and known probe geometry support container-scale coverage while
estimating candidate-object positions, contours, and heights. The robot then
sets down the probe and uses the bilateral visuotactile fingertips to acquire
structured local observations. A training-free, coarse-shape-conditioned
matcher compares these observations directly with the given 3-D target model
and drives recognition and grasping. In this way, one gripper connects
large-scale force-based search to high-resolution haptic recognition while
retaining its end-effector function for target retrieval.

The main contributions of this work are as follows.
\begin{itemize}
    \item We present \textbf{GLoTouch}, a global-to-local haptic perception
    and manipulation framework that completes container-scale force-based
    search and object-scale tactile recognition with the same parallel
    gripper in dark or low-light scenes where external vision is
    unreliable.
    \item Globally, probe-mediated force exploration estimates candidate
    object centers, contours, and heights; locally, a training-free
    finite-window STL matcher directly identifies the target.
    \item In simulation and real-robot experiments, end-to-end retrieval
    succeeds at 84.0\% and 76.0\% in five-object containers, validating
    tactile-only target retrieval.
\end{itemize}

\section{RELATED WORK}
\label{sec:related_work}

\subsection{Tactile Object Search and Retrieval}

When darkness or low illumination makes external vision unreliable, active
touch provides an alternative route to object localization, discrimination,
and target search
\cite{petrovskaya2011global,kaboli2019tactile}. TactoFind
\cite{pai2023tactofind} uses a three-fingered hand with GelSight 360 sensors
to perform grid-based vertical probes and learned tactile representations for
multi-object retrieval. TACTFUL \cite{kamtikar2026tactful} employs
reinforcement learning on a multi-fingered robot to balance workspace coverage
and local surface refinement. DexTouch \cite{lee2024dextouch} similarly
demonstrates that learned tactile policies can coordinate object seeking and
manipulation. Closest to our setting, STUCCO \cite{zhong2022soft} retrieves
a known-shape object without vision using a parallel-jaw gripper equipped
with two soft-bubble tactile sensors, combining particle-filter state
estimation with contact data association to track objects that repeated
contacts move.

These systems demonstrate that multi-object tactile retrieval without
external vision is feasible. TactoFind, TACTFUL, and DexTouch rely on
multi-fingered dexterous hands and learned tactile policies that may require
substantial real-robot data. STUCCO shares our parallel-gripper hardware,
but maintains explicit pose tracking and contact association as objects are
pushed and rotated. In contrast, GLoTouch keeps the simple gripper while
changing both components: a passive probe temporarily held by the gripper
extends contact sensing to container-scale, coverage-driven search, and a
training-free finite-window matcher compares structured bilateral tactile
observations directly with the target STL instead of fusing tracked poses.
One gripper therefore switches between probe-mediated exploration, local
recognition, and grasping in dark or low-light settings without learned
policies or per-object state estimation.

\subsection{Tool-Mediated Exploration and Extrinsic Contact Sensing}

Without external vision, the key bottleneck in global search is workspace
coverage and remote contact sensing. Slender tools and compliant whiskers can
reduce the swept cross section in an unknown workspace and extend contact
sensing beyond locations directly reachable by the end effector
\cite{fox2012tactile}. Xiao \emph{et al.}
\cite{xiao2022active} combine compliant tactile whiskers with
uncertainty-driven path planning for multi-object exploration; Zhao
\emph{et al.} \cite{zhao2024unknown} mount a multi-point tactile sensor at
the tip of a tool rod and manipulate unknown objects in confined spaces.
These studies show the value of tool-mediated contact, but the former
requires a dedicated whisker mechanism and the latter an instrumented tool
tip.

A related body of work senses remote environmental contact from the response
generated at the interface between a robot hand and a grasped object,
including contact-point estimation for uncalibrated tools
\cite{karayiannidis2014online}, transient contact detection via fingertip
tactile signals \cite{molchanov2016contact}, extrinsic contact estimation
combining distributed tactile sensing with kinematic constraints
\cite{ma2021extrinsic}, and joint estimation of grasped-object motion and
contact state \cite{kim2023simultaneous}. Contact SLAM \cite{wang2025contact}
further combines bilateral visuotactile measurements with a known tool model
for probabilistic inference.

These studies establish important foundations for sensing extrinsic contact
through a grasped tool. GLoTouch uses an uninstrumented, passive
spherical-tipped probe solely for global discovery and coarse geometric
estimation; the same gripper then performs high-resolution local tactile
recognition and target grasping.

\subsection{Active Tactile Geometry Acquisition and Object Recognition}

After global candidate localization, local recognition must still recover
discriminative object geometry within a limited contact region. Tactile
geometry has long been used for object mapping, recognition, and
localization \cite{pezzementi2011object,bauza2019tactile}. Because a single
tactile observation covers only a limited surface region, active tactile
recognition must select informative contact actions and aggregate local
observations over time. Prior methods have explored object contours,
probabilistic surface models, and continuous query path optimization
\cite{martinez2013active,yi2016active,driess2017active}, or construct object
models from active contacts
\cite{jamali2016active,fleer2020learning}. TANDEM3D \cite{xu2023tandem3d}
jointly learns exploration and recognition modules to improve object
discrimination; AcTExplore \cite{shahidzadeh2024actexplore} uses
reinforcement learning to improve surface coverage and reconstruction
completeness. FingerSLAM \cite{zhao2023fingerslam} jointly recovers object
location and geometry through closed-loop state estimation; TouchSDF
\cite{comi2024touchsdf} completes a 3-D surface using a learned shape prior.
SimPLE \cite{bauza2024simple} demonstrates that high-resolution visuotactile
feedback can support precise object localization and placement. These methods
primarily address active view selection, observation fusion, or complete
reconstruction, and generally assume that the robot already operates near the
object of interest.

For unconstrained objects in a container, repeated contact and regrasping
introduce unknown translations and rotations between tactile observations
\cite{koval2015pose,suresh2021tactile}. Directly fusing all tactile point
clouds in a fixed world frame would introduce additional state-estimation
and data-association problems. GLoTouch does not aim for complete
reconstruction; it selects structured local actions according to the coarse
contour estimated globally, extracts local tactile geometry and jaw support
widths, and compares them directly with the geometry generated from the
provided target model.

\section{PROBLEM FORMULATION AND SYSTEM OVERVIEW}
\label{sec:system_overview}

\begin{figure}[t]
    \centering
    \includegraphics[width=\linewidth]{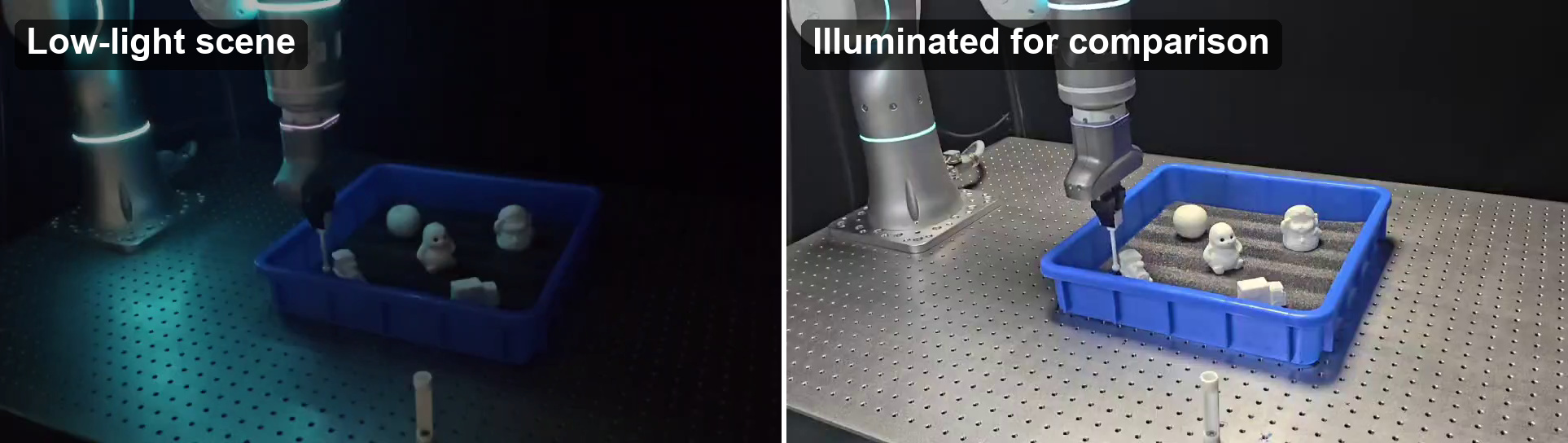}
    \caption{
        Task lighting setting: deployment is in darkness or low light;
        photographs are captured under temporary illumination for
        documentation only. The controller uses no external visual input.
    }
    \label{fig:lighting_conditions}
\end{figure}

\subsection{Task Formulation}

The intended deployment setting is a dark or low-light container in which
external RGB cameras cannot reliably recover object identity, pose, or
geometry, as illustrated in Fig.~\ref{fig:lighting_conditions}. The robot
therefore receives no external camera observation during
control; its sensing is restricted to proprioception, wrist force--torque
measurements, fingertip tactile imaging, and known model geometry. For
experimental documentation, the photographs shown in this paper are recorded
with temporary illumination, but this illumination is not used by the
controller.

Let $\mathcal{B}\subset\mathbb{R}^{3}$ denote the interior workspace of a
container and let $\mathcal{O}=\{O_i\}_{i=1}^{N}$ denote the physical objects
inside it. Their planar locations, poses, and geometries are initially unknown
to the robot, although the scene cardinality $N$ is known. The input is a
target geometry $\mathcal{M}^{\star}$ represented
as an STL triangle mesh. The task is to explore and locate the object
$O^{\star}\in\mathcal{O}$ corresponding to $\mathcal{M}^{\star}$, identify
it, and grasp and retrieve it from the container using touch alone.

We assume that the container boundary and robot-to-container registration are
known, the objects rest on the container floor, and each object can be reached
from above. The objects may translate or rotate when contacted. Success
requires the retrieved physical object to match the provided target mesh.

\subsection{System Setup}

The system consists of a seven-degree-of-freedom Flexiv robot arm, wrist
six-axis force--torque sensing, and an Xense parallel gripper. The gripper
can pick up a probe with no embedded sensor;
environmental contact at its tip is inferred from wrist wrench measurements
and robot kinematics. After global exploration, the gripper places the probe
at a designated holder
and becomes available for direct bilateral contact and grasping.

At time $t$, the global stage observes the TCP transform
$\mathbf{T}^{B}_{E,t}\in SE(3)$ and the wrist force--torque wrench
$\mathbf{w}_{t}=[\mathbf{f}_{t}^{\mathsf T},
\boldsymbol{\tau}_{t}^{\mathsf T}]^{\mathsf T}$, used for contact detection
and safety stopping. The local stage observes the left and right tactile
depth maps $D^{L}_{t}$ and $D^{R}_{t}$ and the parallel-jaw aperture $g_t$;
the stable tactile depth and aperture measurements provide the local
geometric observation used for model matching.

\subsection{Global-to-Local Pipeline}

Figure~\ref{fig:glotouch_pipeline} summarizes the two scales of GLoTouch. At
the container scale, the held probe moves
at a fixed lateral sensing height and performs coverage-driven search;
after detecting a contact, multi-directional probing estimates candidate
positions, contours, and heights. At the object scale, the gripper sets down
the probe and sequentially contacts each candidate to acquire structured
local haptic observations. After all candidates are collected, the
observations are matched against the target model and ranked by matching loss;
the object with the lowest matching loss is then grasped and retrieved.

\section{PROBE-MEDIATED GLOBAL HAPTIC EXPLORATION}
\label{sec:global_exploration}

\subsection{Contact Detection and Active Retreat}

\begin{figure}[t]
    \centering
    \begin{minipage}[t]{0.49\columnwidth}
        \vspace{0pt}
        \centering
        \includegraphics[width=\linewidth]{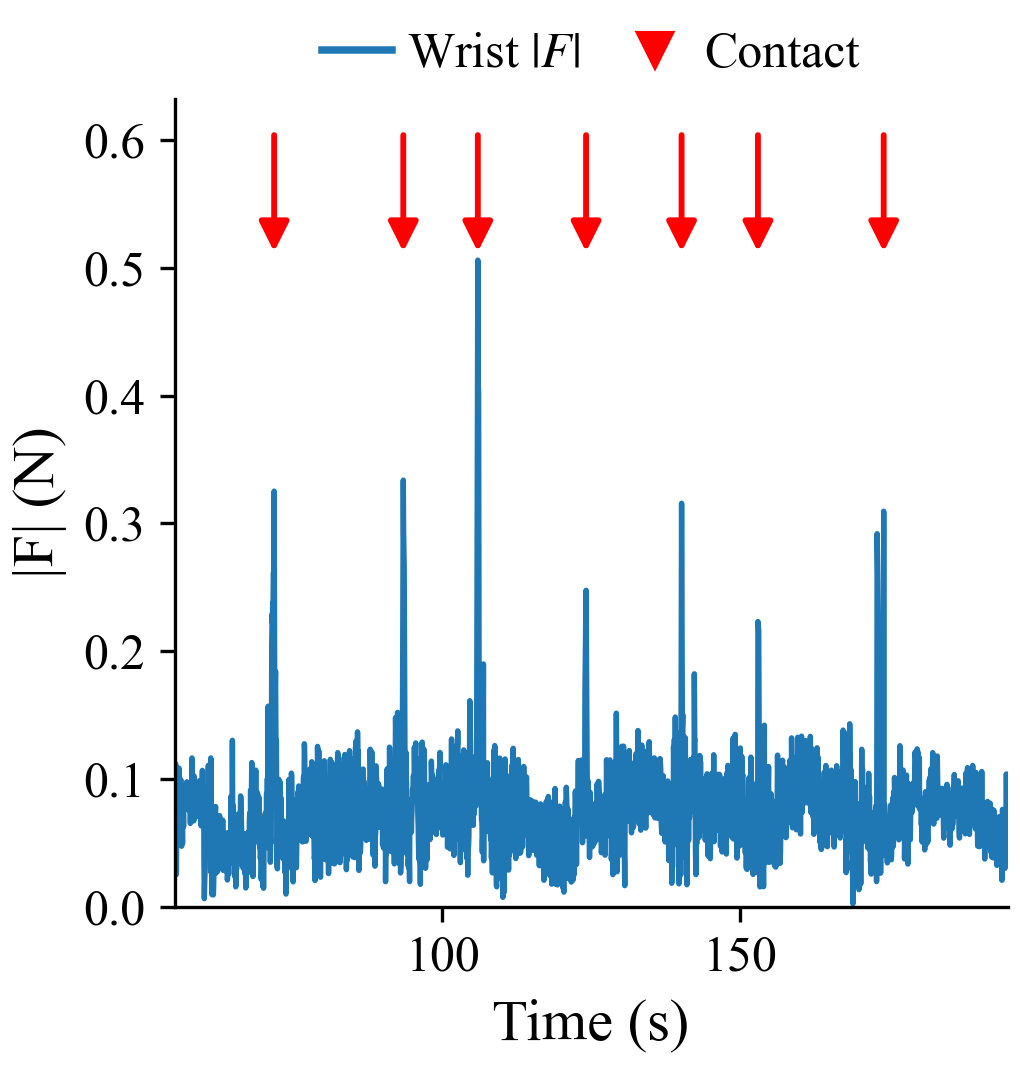}

        \vspace{0.3em}

        {\small
        \textbf{(a)} Wrist force response.}
    \end{minipage}
    \hfill
    \begin{minipage}[t]{0.49\columnwidth}
        \vspace{0pt}
        \centering
        \includegraphics[width=\linewidth]{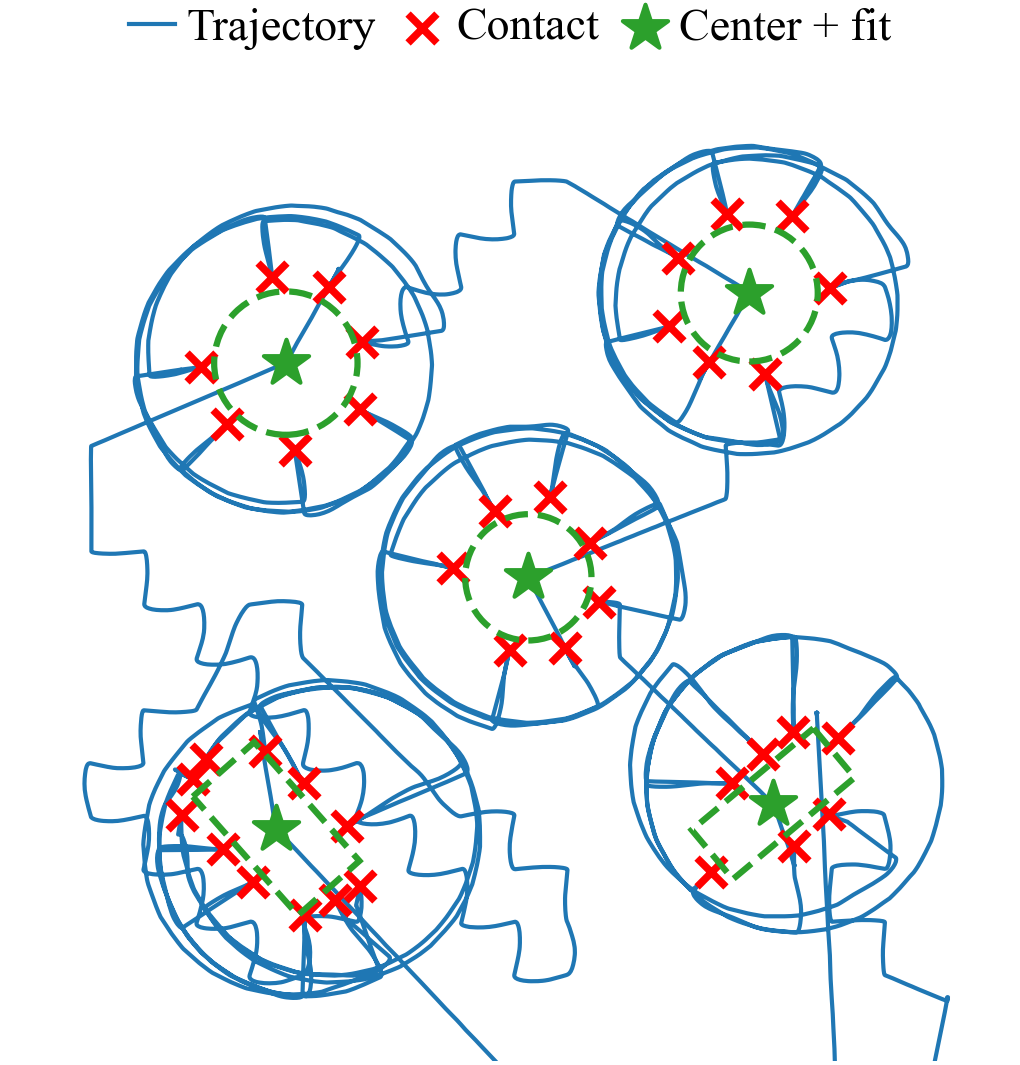}

        \vspace{0.3em}

        {\small
        \textbf{(b)} Top-down exploration map.}
    \end{minipage}
    \caption{
        Global exploration and localization.
        \textbf{(a)} Wrist force magnitude; red arrows mark
        contact-triggered stops.
        \textbf{(b)} Exploration trajectory, contact points, estimated
        object centers, and fitted contours.
    }
    \label{fig:global_exploration_results}
\end{figure}

Figure~\ref{fig:global_exploration_results} shows a representative force
response and the top-down exploration map. Before
exploration, the wrist force--torque sensor is zeroed with the tool
unloaded. A stationary interval provides a robust wrench baseline
$\bar{\mathbf{w}}$, and the online contact signal is the low-pass-filtered
residual
\begin{equation}
    \Delta\mathbf{w}_{t}
    = \operatorname{LPF}(\mathbf{w}_{t}-\bar{\mathbf{w}}).
    \label{eq:wrench_residual}
\end{equation}
Here $\mathbf{w}_{t}$ is the measured wrist wrench and $\operatorname{LPF}$
denotes low-pass filtering. For a commanded planar motion direction
$\mathbf{u}_t$, a normal contact is
declared when the opposing directional force and the total force or moment
exceed their soft thresholds for a fixed number of consecutive samples. A
separate set of absolute and residual wrench limits triggers an immediate
hard stop. Soft thresholds preserve sensing sensitivity; hard stops retain
an independent safety envelope.

Upon detecting a normal contact, motion stops immediately and the onset pose
and wrench are retained. The probe retreats a short distance opposite to its
approach; contour tracking begins after wrench and Cartesian velocity return
to a stable interval.

\subsection{Tool-Geometry-Aware Contact Localization}

Let $\mathcal{P}$ be the known planar cross section of the held probe at the
object-height band and let $\mathbf{x}_{E,t}\in\mathbb{R}^{2}$ be the planar
TCP position at contact onset. The measured reaction direction in the base
frame is
\begin{equation}
    \mathbf{r}_{t}=\frac{\Delta\mathbf{f}^{B}_{t,xy}}
    {\lVert\Delta\mathbf{f}^{B}_{t,xy}\rVert_2}.
\end{equation}
where $\Delta\mathbf{f}^{B}_{t,xy}$ is the filtered planar force residual
in the robot base frame.
Global search may contact a face or a corner, so the object-facing direction
combines the known approach and the opposing reaction,
\begin{equation}
    \mathbf{d}_{t}=
    \frac{\mathbf{u}_{t}-\mathbf{r}_{t}}
    {\lVert\mathbf{u}_{t}-\mathbf{r}_{t}\rVert_2}.
    \label{eq:contact_direction}
\end{equation}
For an explicitly radial trace probe, $\mathbf{d}_{t}=\mathbf{u}_{t}$ after
verifying directional consistency.

The support distance of the probe in direction $\mathbf{d}_{t}$ is
\begin{equation}
    h_{\mathcal{P}}(\mathbf{d}_{t})
    =\max_{\mathbf{p}\in\mathcal{P}}\mathbf{d}_{t}^{\mathsf T}
    \mathbf{R}^{B}_{E,t}\mathbf{p}.
    \label{eq:support_function}
\end{equation}
where $\mathbf{p}$ is a probe-section point and $\mathbf{R}^{B}_{E,t}$
rotates the probe section into the robot base frame.
The environmental contact is represented by the support-plane intercept
\begin{equation}
    \hat{\mathbf{p}}_{t}=
    \mathbf{x}_{E,t}+h_{\mathcal{P}}(\mathbf{d}_{t})\mathbf{d}_{t}.
    \label{eq:contact_point}
\end{equation}
An initial object-center hypothesis is placed a nominal radius inside the
boundary and refined with each~\mbox{contact}.

\subsection{Coverage-Driven Object Search}

The search policy maintains samples $\mathcal{E}$ along the TCP's executed
planar path. Frontier candidates are sampled uniformly within the permitted
container boundary; a candidate $\mathbf{q}$ remains unexplored when
\begin{equation}
    \min_{\mathbf{e}\in\mathcal{E}}
    \lVert\mathbf{q}-\mathbf{e}\rVert_2>r_{\mathrm{cov}}.
\end{equation}
where $\mathbf{e}$ indexes executed-path samples and $r_{\mathrm{cov}}$ is
the coverage radius.
Traced object centers are registered in the policy's exclusion list;
duplicate contacts within a traced object's envelope are skipped and global
search resumes immediately. The nearest unexplored frontier is selected,
producing continuous Cartesian segments through the container. A segment
without contact is added to $\mathcal{E}$; a segment with contact switches
to local boundary tracing. Search resumes after the object is
characterized, continuing until all expected objects are found or
the search space is exhausted.

\subsection{Orbit-and-Probe Contour Estimation}

From the current center estimate $\hat{\mathbf{c}}$, the robot moves outward
to a clearance orbit, follows the orbit in bounded angular increments, and
approaches radially toward the center. The orbital portion avoids chord
paths through the object; the radial portion produces boundary observations
with known approach directions. After each accepted contact, the robot
retreats and updates the center before planning the next direction,
prioritizing the largest uncovered angular interval.

For a round-like candidate, planar contact points
$\{\hat{\mathbf{p}}_k\}_{k=1}^{K}$ are fit with a circle by minimizing
\begin{equation}
    \min_{\mathbf{c},\rho}
    \sum_{k=1}^{K}
    \rho_{\mathrm{SL1}}
    \left(\lVert\hat{\mathbf{p}}_k-\mathbf{c}\rVert_2-\rho\right),
    \label{eq:circle_fit}
\end{equation}
where $K$ is the number of accepted contacts, $\mathbf{c}$ and $\rho$ are
the fitted circle center and radius, and $\rho_{\mathrm{SL1}}$ is a
soft-$\ell_1$ robust loss. When the circle
residual remains high after the minimum number of contacts, additional
directions are collected and an oriented rectangle is fitted using
point-to-nearest-edge residuals. The selected primitive supplies the center,
coarse contour class, scale, and, for a box-like shape, its planar
orientation.

Finally, the probe is positioned above the fitted center and moves downward
under the same guarded contact detector. The object-height estimate is the
top-contact height minus the calibrated floor height. The candidate record
passed to the local stage is
$\mathcal{C}_i=(\hat{\mathbf{c}}_i,\hat{H}_i,s_i,
\hat{\theta}_i,k_i,q_i)$, where the entries denote center, height, scale,
planar orientation, coarse contour class
$k_i\in\{\mathrm{round},\mathrm{box}\}$, and local-fit quality.

After all objects are localized, the probe is returned to its holder and the
gripper transitions to the local stage.

\subsection{Contact-Mode Validation}

\begin{figure}[t]
    \centering
    \newsavebox{\gtbox}
    \sbox{\gtbox}{\includegraphics[width=0.52\columnwidth]{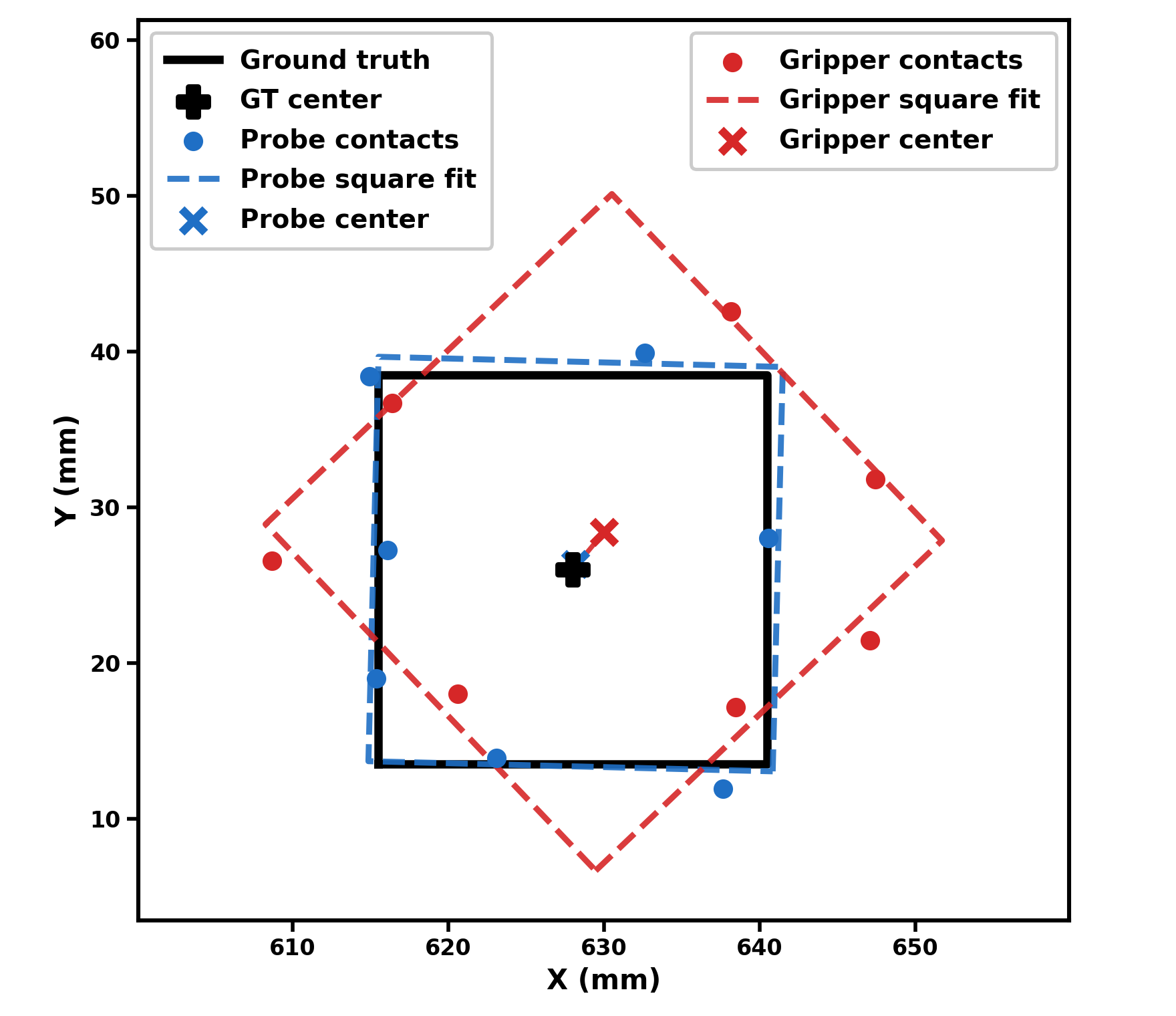}}
    \newlength{\gtheight}
    \setlength{\gtheight}{\ht\gtbox}

    \begin{minipage}[b]{0.52\columnwidth}
        \centering
        \usebox{\gtbox}

        \vspace{0.3em}

        {\small
        \textbf{(a)} Contacts and center estimates.}
    \end{minipage}
    \hfill
    \begin{minipage}[b]{0.46\columnwidth}
        \centering
        \parbox[c][\gtheight][c]{\linewidth}{%
            \centering
            {\scriptsize
            \renewcommand{\arraystretch}{1.5}
            \setlength{\tabcolsep}{2.5pt}
            \begin{tabular}{l|cc}
                \hline
                Metric & Probe & Gripper \\
                \hline
                Center error & \textbf{0.37} & 3.15 \\
                Contact std & \textbf{0.84} & 4.81 \\
                Contact RMS & \textbf{0.87} & 5.08 \\
                Max contact & \textbf{1.58} & 6.93 \\
                Side error & \textbf{0.95} & 5.67 \\
                \hline
            \end{tabular}
            }%
        }

        \vspace{0.3em}

        {\small
        \textbf{(b)} Errors (mm).}
    \end{minipage}
    \caption{
        Contact-mode validation on a calibration square.
        \textbf{(a)} Ground truth, contacts, square fits, and centers.
        \textbf{(b)} Errors from the true contour~(mm).
    }
    \label{fig:center_error}
\end{figure}

The global stage relies on probe-mediated contact rather than direct
gripper contact. To validate this design choice, both modes execute
the same orbit-and-probe procedure on a calibration square with
known pose, and the fitted squares are compared against the ground
truth (Figure~\ref{fig:center_error}).

Probe-mediated tracing fits the contour well, while the bare gripper
produces noticeably larger center error and contact scatter, together
with a misaligned contour fit. This stems from contact
localization: the gripper's large envelope makes face contact and the
effective contact point uncertain, validating the necessity of
probe-mediated~\mbox{tracing}.

\section{LOCAL HAPTIC RECOGNITION AND TARGET RETRIEVAL}
\label{sec:local_recognition}

\subsection{Stable Tactile Acquisition and Quality Assessment}

After global exploration, the robot returns the probe and moves the open
gripper above a candidate. At each height layer, the jaws perform a
force-limited closure. A layer is accepted only when the aperture reaches a
stable plateau and TCP motion is small; the opening is recorded as $g_{k,h}$,
where $k$ indexes a closure sequence and $h$ a height layer. Unstable layers
are skipped.

For each tactile side $s\in\{L,R\}$ and TouchSet $j$, temporal median filtering
gives a depth map $D_j^s(u,v)$. Indentations above a small threshold form a
contact mask, and only sufficiently large connected support patches are
retained. Fitting a plane to such a patch yields the relative nonplanarity
\begin{equation}
    \rho_{j,k}^s =
    \frac{\operatorname{RMS}(D_j^s-\widehat D_j^s)}
         {q_{90}(D_j^s)},
    \label{eq:relative_nonplanarity}
\end{equation}
where $\widehat D_j^s$ is the fitted plane and $q_{90}$ is the
90th-percentile indentation. The sequence-level $\rho_k$ is the median over
valid left/right patches, and $q_{D,k}$ is the fraction of valid tactile
patches. Bilateral layers receive the highest weight, unilateral layers are
down-weighted, and invalid layers are excluded.

\subsection{Target-Conditioned Finite-Window Matching}

The global contour selects one of two acquisition modes: round-like candidates
are sampled at three independent yaws, whereas rectangular candidates are
sampled along three parallel longitudinal tracks. Each candidate therefore
yields a TouchSet $D_j$ containing three multi-height sequences.

The reference open-TCP positions $\mathbf{t}_k$ and rotations $R_k$ infer an
action graph. SVD of the centered positions gives longitudinal coordinates
$a_k$. With $\Delta t=\max_k a_k-\min_k a_k$ and
$\Delta R=\max_{m<n}d(R_m,R_n)$, where $d(\cdot,\cdot)$ is relative rotation,
the soft coupling probability is
\begin{equation}
    p_c =
    \sqrt{
    \sigma\!\left(3\log\frac{\max(\Delta t,\epsilon_t)}{\tau_t}\right)
    \sigma\!\left(-3\log\frac{\max(\Delta R,\epsilon_R)}{\tau_R}\right)
    },
    \label{eq:coupled_probability}
\end{equation}
where $\sigma(\cdot)$ is a logistic function, $\epsilon_t$ and $\epsilon_R$
are positive floors, and $\tau_t,\tau_R$ are reference scales.

For the target STL $\mathcal{M}$, let $a_M$ denote its horizontal
bounding-box anisotropy. For scale $s$, relative yaw $\theta$, and tactile
window $\mathcal{W}(l,z)$, the model predicts support width
\begin{equation}
    W_{\mathcal{M}}(l,z;s,\theta)=
    \max_{\mathbf{v}\in\mathcal{W}(l,z)}c(\mathbf{v})
    -
    \min_{\mathbf{v}\in\mathcal{W}(l,z)}c(\mathbf{v}),
    \label{eq:finite_window_width}
\end{equation}
where $l$ and $c(\mathbf{v})$ are longitudinal and closing-direction
coordinates. Surface-normal dispersion $d_M$ is calibrated to the
sensor-domain prediction $\rho_M$.

For nuisance parameters $s$, $\theta$, window shift $\delta$, scan-top offset
$z_0$, scan reflection, and aperture bias $b$, the predicted opening is
\begin{equation}
    \widehat y_{k,h}=
    W_{\mathcal{M}}(a_k+\delta,z_{k,h}+z_0;s,\theta)+b.
    \label{eq:predicted_width}
\end{equation}
All residuals use a weighted Huber RMS~\cite{huber1992robust}. The width loss
combines absolute, centered, vertical-difference, and cross-action residuals,
plus coverage and aperture-bias~\mbox{penalties}:
\begin{align}
    \mathcal{G}={}&
    w_aH_a(y-\widehat y;w)
    +w_cH_c(\tilde y-\tilde{\widehat y};w)\nonumber\\
    &+w_vH_v(\Delta_z y-\Delta_z\widehat y;w)\nonumber\\
    &+w_xH_x(\Delta_a y-\Delta_a\widehat y;w)
    +\lambda_C(1-C)
    +\lambda_b\frac{|b|}{b_0},
    \label{eq:width_sensor_loss}
\end{align}
where $H_{a,c,v,x}$ are weighted Huber RMS operators, $w_{a,c,v,x}$ are loss
weights, $w$ is observation evidence, $C$ is the finite-support fraction, and
$b_0$ normalizes the bias. A coupled branch shares one model pose across the
three sequences; an independent branch fits each sequence separately and fuses
the losses by median. Each branch objective adds a scale prior
$\lambda_s|\log s|$, whose weight increases with model anisotropy $a_M$. The
width loss is their soft mixture
\begin{equation}
    \mathcal{L}_{\mathrm{width}}
    =
    p_c\mathcal{L}_{\mathrm{width}}^{\mathrm{coupled}}
    +(1-p_c)\mathcal{L}_{\mathrm{width}}^{\mathrm{independent}}.
    \label{eq:soft_action_graph}
\end{equation}

The local-depth term compares $\rho_k$ with $\rho_M$ and weights the residual
by model surface information. With $I_M=d_M/(d_M+\epsilon)$, the depth loss is
\begin{equation}
    \mathcal{L}_{\mathrm{depth}}
    =\!
    \operatorname{median}_k
    \log\!\left[
    1+
    \left(\frac{\rho_k-\rho_M}{\sigma_D}\right)^2
    \right]\!
    \operatorname{mean}_k(q_{D,k}I_M)
    \label{eq:depth_loss}
\end{equation}
Here $\sigma_D$ is the depth-noise scale. Its weight
$\lambda_D=\lambda_{D,0}/(1+\gamma a_M)$ decreases with $a_M$ because the
width profile is already informative for anisotropic targets. A
collection-mode consistency term
\begin{equation}
    \mathcal{L}_{\mathrm{cm}}
    =
    \lambda_{\mathrm{cm}}a_M^2(1-p_c)
    \label{eq:collection_mode}
\end{equation}
penalizes the mismatch between anisotropic models and independently fitted
sequences. The final matching loss is
\begin{equation}
    \mathcal{L}_{\mathrm{match}}(D_j\mid\mathcal{M})=
    \mathcal{L}_{\mathrm{width}}
    +\lambda_D\mathcal{L}_{\mathrm{depth}}
    +\mathcal{L}_{\mathrm{cm}}.
    \label{eq:matching_loss}
\end{equation}

\subsection{Ranking and Target Grasping}

For a fixed target mesh $\mathcal{M}$, the matcher ranks all touched physical
TouchSets by $\mathcal{L}_{\mathrm{match}}(D_j\mid\mathcal{M})$. A decision is
accepted only when the best candidate has sufficient first--second loss
margin, absolute fit quality, and tactile evidence. The grasp center is
corrected by the local contact center, the rectangular orientation is
initialized from the global major direction, and the gripper closes under the
same force limit before lifting the selected object.

\begin{figure}[!t]
    \centering
    \includegraphics[width=\linewidth]{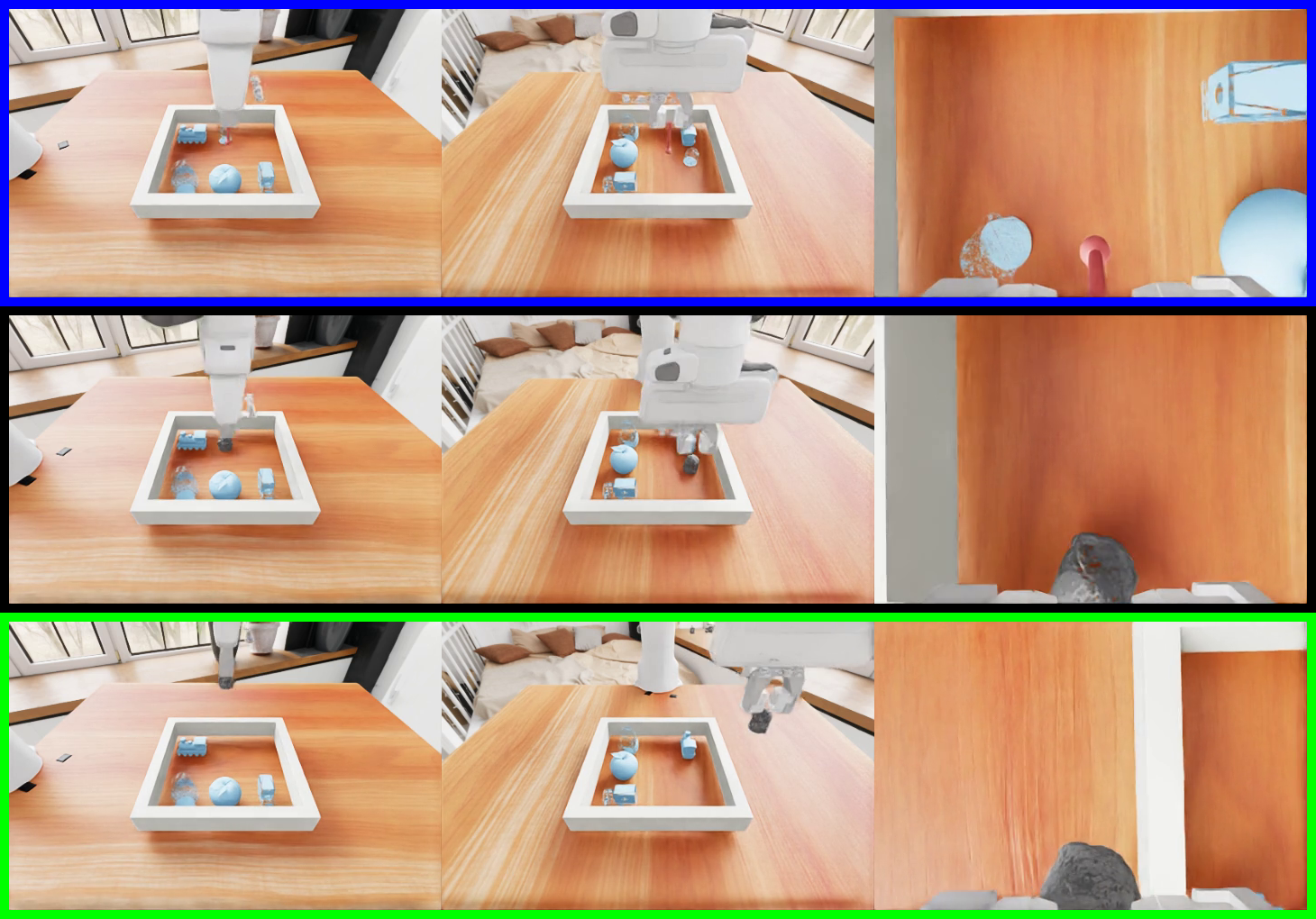}
    \caption{
        UniVTAC simulation sequence.
        \textcolor{blue}{Blue}: probe-mediated global exploration.
        \textcolor{black}{Black}: tactile acquisition.
        \textcolor{green!60!black}{Green}: grasping.
    }
    \label{fig:simulation_environment}
\end{figure}

\section{EXPERIMENTS}
\label{sec:experiments}

\subsection{Simulation Benchmark}
\label{sec:simulation}

We port the full exploration--retrieval pipeline to UniVTAC
\cite{chen2026univtac}. STL objects are converted to UIPC bodies to obtain
deformable fingertip contact, but these bodies are outside the PhysX
contact-reporting pipeline, whereas blind exploration relies on
collision-triggered stopping and orbit tracing. During exploration, each
object therefore has a co-located PhysX proxy with identical geometry and pose
registered to the robot contact pipeline. After a candidate is localized, the
proxy is removed and the UIPC body takes over guarded descent and removal at
the same task state. The probe is modeled as a rigid body with a
ContactSensor, so remote collisions enter the controller as wrench readings,
consistent with real-robot inference from TCP wrench.
Fig.~\ref{fig:simulation_environment} shows the simulation environment.

Table~\ref{tab:simulation} reports per-target end-to-end retrieval in the
five-object simulation benchmark; each target is evaluated in 20 randomized
trials, 100 in total. GLoTouch retrieves 84 of the 100 targets (84.0\%).
Apple is the most reliably retrieved target (20/20, 100.0\%), while Train
is the most difficult (14/20, 70.0\%). Failures mainly arise from exploration
runs that do not discover all candidates, occasional simulator runtime
failures, and the largest recognition confusion, between Train and Truck.

We additionally test whether global search is necessary using a
no-global-search/random-descent baseline on 100 randomized five-object
layouts. The baseline has no contact map or frontier selection: it samples a
random descent site, moves there without sweep detection, descends vertically,
and uses the same local centering procedure under the same budget. As shown in
Table~\ref{tab:global_search_ablation}, random descent discovers fewer objects
at every budget and completes all five discoveries in only 30\% of layouts with
a 5~m budget, compared with 55\% for coverage-driven search. This gap shows
that random descents can find isolated objects but structured global search
improves complete scene discovery.

\begin{table}[!t]
\centering
\caption{Global-search ablation on 100 randomized five-object layouts. Discovered values are ordered as 1/3/5~m budgets; complete discovery is at 5~m.}
\label{tab:global_search_ablation}
\renewcommand{\arraystretch}{1.15}
\begin{tabular}{lcc}
\hline
Method & Mean discovered (/5) & Complete at 5~m (\%) \\
\hline
Random descent & 1.23/3.17/4.02 & 30.0 \\
Global search & 1.35/3.30/4.44 & 55.0 \\
\hline
\end{tabular}
\end{table}

\begin{table}[!t]
\centering
\caption{Simulation retrieval by target in five-object scenes. Each target is
evaluated in 20 trials.}
\label{tab:simulation}
\small
\setlength{\tabcolsep}{3pt}
\renewcommand{\arraystretch}{1.15}
\begin{tabular}{lcccccc}
\hline
Target & Wukong & Nailong & Apple & Train & Truck & Total \\
\hline
Successes & 16/20 & 19/20 & 20/20 & 14/20 & 15/20 & 84/100 \\
\hline
\end{tabular}
\end{table}

\subsection{Local Tactile Matching Validation}
\label{sec:local_matching_validation}

Before evaluating the full pipeline on hardware, we validate the local matcher
as an independent component. This isolates model-conditioned recognition from
global search and grasping. The evaluation uses 60 target/phase samples from
five physical objects. Each trial compares a target mesh against a TouchSet
assembled from three acquisition sequences of the same physical object.
Different phases reuse different sequence subsets, so the protocol measures
target-conditioned matching under repeated contact. With parameters frozen
before these trials, the full matcher achieves 90.0\% Top-1 accuracy (54/60)
with a mean first--second loss margin of 0.402. Width/action matching alone
obtains 68.3\% (41/60) and 0.267. Adding local depth changes the result to
78.3\% (47/60) and 0.343; adding collection-mode consistency gives the
full-method result. Table~\ref{tab:loss_matrix} shows that correct pairings
receive distinctly lower matching loss: rows are target STLs, columns are
physical objects, and each column averages the losses over that object's
phase TouchSets. Specifically, the matrix should be read row by row: each row
fixes a specified target STL and evaluates it against TouchSets from all
physical objects. Within a row, the lowest-loss column is the recognition
decision, and the diagonal corresponds to the correct physical object.
Therefore, differences among cells in a row measure the discrimination margin
for that target, whereas a column only averages the same physical object's
TouchSet losses against different STLs and is not itself a separate recognition
decision. Table~\ref{tab:ablation} reports the progressive~\mbox{ablation}.

\begin{table}[!t]
\centering
\footnotesize
\setlength{\tabcolsep}{3pt}
\caption{Mean matching-loss matrix under the full matcher. Rows: target STLs;
columns: physical objects. Diagonal entries: correct pairings.}
\label{tab:loss_matrix}
\renewcommand{\arraystretch}{1.15}
\begin{tabular}{l|ccccc}
\hline
STL \textbackslash{} Object & Nailong & Wukong & Train & Apple & Truck \\
\hline
Nailong & \textbf{0.27} & 0.48 & 2.40 & 0.65 & 1.69 \\
Wukong & 1.02 & \textbf{0.54} & 2.27 & 1.47 & 1.85 \\
Train & 1.87 & 1.33 & \textbf{0.78} & 1.37 & 0.93 \\
Apple & 1.59 & 0.89 & 2.96 & \textbf{0.08} & 2.24 \\
Truck & 1.91 & 2.02 & 2.05 & 1.25 & \textbf{0.90} \\
\hline
\end{tabular}
\end{table}

\begin{table}[!t]
\centering
\caption{Progressive ablation on the five-object development set.
Margin is the mean first--second loss difference under each method's own
loss scale and is not directly comparable across rows.}
\label{tab:ablation}
\renewcommand{\arraystretch}{1.15}
\begin{tabular}{lcc}
\hline
Method & Top-1 (\%) & Margin \\
\hline
Width/action only & 68.3 & 0.267 \\
\,+ Local depth & 78.3 & 0.343 \\
\,+ Collection-mode prior (Full) & \textbf{90.0} & \textbf{0.402} \\
\hline
\end{tabular}
\end{table}

\subsection{Real-Robot Experiments}
\label{sec:real_robot_experiments}

\begin{figure}[!t]
    \centering
    \includegraphics[width=\linewidth]{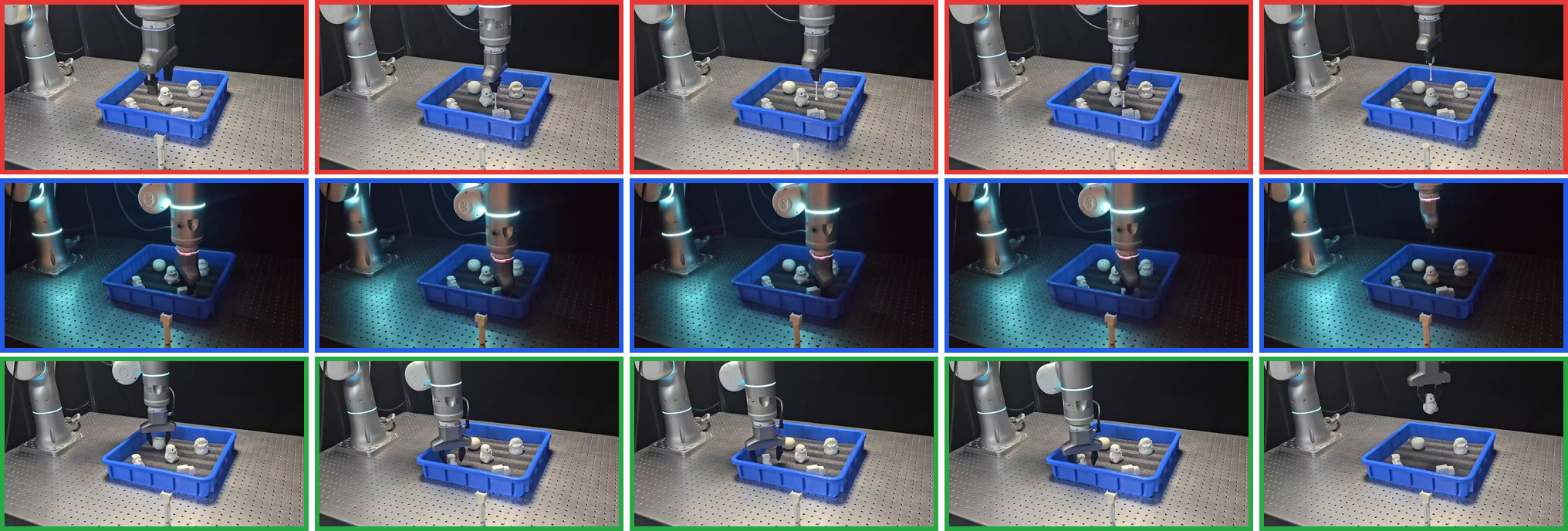}
    \caption{
        Real-robot five-object sequence.
        \textcolor{red!80!black}{Red}: global probe search.
        \textcolor{blue!70!black}{Blue}: probe return and local recognition.
        \textcolor{green!60!black}{Green}: target retrieval.
        Dark frames indicate a temporary lights-off period; the tactile-driven
        pipeline continues to execute.
    }
    \label{fig:real_sequence}
\end{figure}

Real experiments use the Flexiv and Xense setup described in
Sec.~\ref{sec:system_overview}. The robot receives the target STL mesh and
known scene cardinality, but not target identity, object poses, or contact
correspondences. Five-object scenes contain one target and four
distractors. Object poses and
yaws are sampled randomly without collision. The object set contains five
3-D-printed objects (Wukong, Nailong, Apple, Train, and Truck), as visible
in Fig.~\ref{fig:real_sequence}. All contact, acquisition, and matching
parameters are frozen before any reported trial and are never adjusted per
object or per scene. External measurements are used only for ground-truth
annotation and~visualization.

End-to-end success requires discovering the target, identifying it correctly,
establishing a grasp, and lifting it clear of the container without external
visual input or human correction. In the five-object benchmark, each target is
evaluated in 10 randomized trials, giving 50 real-robot trials in total.
GLoTouch retrieves the specified target in 38 trials, yielding a 76.0\%
end-to-end success rate. Table~\ref{tab:real_object_results} reports the
per-target results. Apple achieves 100.0\% retrieval, whereas Train is the
most difficult target with a 50.0\% success rate; Wukong, Nailong, and Truck
achieve 70.0--80.0\%.

\begin{table}[!t]
\centering
\caption{Real-robot end-to-end retrieval by target in five-object scenes.
Each target is evaluated in 10 trials.}
\label{tab:real_object_results}
\small
\setlength{\tabcolsep}{3pt}
\renewcommand{\arraystretch}{1.15}
\begin{tabular}{lcccccc}
\hline
Target & Wukong & Nailong & Apple & Train & Truck & Total \\
\hline
Successes & 7/10 & 8/10 & 10/10 & 5/10 & 8/10 & 38/50 \\
\hline
\end{tabular}
\end{table}

Failure analysis shows that the main recognition errors occur between Train
and Truck, and the failure direction is relatively clear: a specified Train
target is grasped as Truck. This outcome is consistent with the component-level
matrix in Table~\ref{tab:loss_matrix}: for the Train target, the Train STL
obtains 0.78 on the Train TouchSet and 0.93 on the Truck TouchSet, yielding
only a 0.15 mean margin, whereas the Truck target obtains 0.90 on the Truck
TouchSet and 2.05 on the Train TouchSet. The other failure mode arises during
physical execution.
Objects can rotate slightly under probe contact during tactile acquisition,
shifting the local observations relative to the target-model acquisition mode.
When grasping Train and Truck, objects can also topple during clamping, so
retrieval fails after reaching the grasping stage. These remaining
failures therefore reflect both the boundary of local geometric discrimination
and the stability of physical manipulation.

Figure~\ref{fig:real_sequence} illustrates a representative five-object run
in three rows. The red row shows the gripper-held probe performing
contact-driven global search in the container. The blue row shows local
bilateral fingertip acquisition after the probe is returned; its darkened
images correspond to the temporary low-light operating condition. The green
row shows the system grasping and lifting the object with the lowest matching
loss. Thus, the tactile-driven pipeline remains active across global search,
recognition, and retrieval despite the illumination change.

\section{CONCLUSION}
\label{sec:conclusion}

We presented GLoTouch, a global-to-local haptic perception and manipulation
framework for retrieving a model-specified object from a multi-object
container without external vision. A parallel gripper temporarily holds a
passive spherical-tipped probe for coverage-driven force-based search,
tool-aware contact localization, and coarse contour and height estimation; it
then sets down the probe and uses the same gripper's bilateral visuotactile
sensors for shape-conditioned, multi-height contact acquisition, which
finite-window support geometry links directly to the target STL without
object-specific training. Experiments show that this global-to-local coupling
is effective: coverage-driven global search completes all discoveries in
55\% of 100 randomized five-object layouts, compared with 30\% for random
descent; the component-level matcher achieves 90.0\% Top-1 accuracy over 60
target/phase samples; and end-to-end retrieval reaches 84.0\% (84/100) in
simulation and 76.0\% (38/50) on hardware in five-object scenes, with
remaining failures concentrated on closely matched Train/Truck geometry and
physical stability during clamping.

Current boundaries point directly to future work. The system currently
assumes a known target STL, top-accessible objects, and a registered
container frame. Within this setting, the local matcher does not yet exploit
the visuotactile RGB channel, and coarse contours use only round and box
primitives. Future work will enrich local cues with the RGB tactile channel,
reduce reliance on known target models through model-free recognition, extend
coarse shape models beyond round and box-like objects, select local touches
online according to matching uncertainty, and ultimately address dense
clutter in which the target's top is not initially accessible. Larger-scale
real-robot validation will accompany these extensions.

\bibliographystyle{IEEEtran}
\bibliography{bibliography/references}

\end{document}